\documentclass[letterpaper, 10pt, conference]{ieeeconf}
\IEEEoverridecommandlockouts    % to override locked commands

\usepackage{amsmath,amssymb,amsfonts}
\usepackage{cite}
\usepackage{graphicx}
\usepackage{textcomp}
\usepackage{xcolor}
\usepackage{bbm}
\let\labelindent\relax
\usepackage{enumitem}
\usepackage{tabularx}
\usepackage{dirtytalk}
\usepackage{multirow}
\usepackage{subcaption}
\usepackage[table]{xcolor} % in preamble

\usepackage{times}
\usepackage{multicol}

\usepackage{graphicx}
\usepackage{bm}
\usepackage[nolist]{acronym}

\usepackage[linesnumbered,ruled]{algorithm2e}

\usepackage{mathtools}
\usepackage{makecell}
\usepackage{array}
\usepackage{dblfloatfix}
\usepackage{diagbox}
\usepackage{etoolbox}\AtBeginEnvironment{algorithmic}{\small}
\usepackage{soul}
\newif\ifshowcomments

\usepackage{placeins} %prevent tables and figures to go after refernece
 
 \showcommentsfalse %to hide comments

\ifshowcomments %jovin added
    \newcommand{\bae}[1]{\hl{[SB: #1]}\protect\color{black}} % Sangjae added, Aug 24, 2021 
    \newcommand{\di}[1]{\hl{[DI: #1]}\protect\color{black}} % David added, Aug 24, 2021
    \newcommand{\ft}[1]{\hl{[FT: #1]}\protect\color{black}} % Faizan
    \newcommand{\jd}[1]{\hl{[JD: #1]}\protect\color{black}} % Jovin
    \newcommand{\edgar}[1]{{\color{red}{LKC: #1}}\protect\color{black}}
    \newcommand{\shrey}[1]{\hl{[SK: #1]}\protect\color{black}}
    \newcommand{\yifan}[1]{\blue{[YIFAN: #1]}\protect\color{black}}
    
\else
    \newcommand{\bae}[1]{}
    \newcommand{\di}[1]{}
    \newcommand{\ft}[1]{}
    \newcommand{\jd}[1]{}
    \newcommand{\edgar}[1]{}
    \newcommand{\shrey}[1]{}
    \newcommand{\yifan}[1]{#1}

\fi

\usepackage{lipsum}
\usepackage[colorinlistoftodos]{todonotes}

\usepackage{amsthm}

\usepackage{bbm} % for mathbb 1
\usepackage[colorlinks = True, linkcolor = blue, citecolor = blue]{hyperref}

\title{\LARGE \bf N3P: Accelerated Automated Parking via a Learning-Based Naturalistic Three-Stage Scheme}
\author{Yifan Xue$^{1,2}$, Toktam Mohammadnejad$^1$, Faizan M Tariq$^1$, Sangjae Bae$^1$, David Isele$^1$, \\
Yosuke Sakamoto$^1$, Nadia Figueroa$^2$, Jovin D'sa$^1$ 
\thanks{$^1$Honda Research Institute (HRI), Mountain View, CA. $^2$ University of Pennsylvania, PA. The work was done during Yifan Xue's internship at HRI.
(Email: \href{mailto:toktam_mohammadnejad@honda-ri.com}{\texttt{toktam\_mohammadnejad@honda-ri.com}}, \href{mailto:jovin_dsa@honda-ri.com}{\texttt{jovin\_dsa@honda-ri.com}}, \& \href{mailto:yifanxue@seas.upenn.edu}{\texttt{yifanxue@seas.upenn.edu}}.}
}

\begin{document}

\maketitle
% \thispagestyle{empty}
% \pagestyle{empty}

%%%%%%%%%%%%%%%%%%%%%%%%%%%%%%%%%%%%%%%%%%%%%%%%%%%%%%%%%%%%%%%%%%%%%%%%%%%%%%%%

\begin{abstract}
Autonomous parking requires efficient path planning that ensures kinematic feasibility and collision avoidance in constrained environments. Hybrid A* is widely used but computationally expensive, while reinforcement learning (RL) methods lack reliability and often struggle with long-horizon geometric constraints, leading to suboptimal trajectories. We present N3P, a fast learning-based three-stage framework for automated parking. %perpendicular and parallel parking. 
By introducing an intermediate preparatory pose and using a learning module to predict it, N3P decomposes the maneuver into simpler subproblems, thereby reducing computational complexity and accelerating path generation. We validate the framework by integrating it with Hybrid A* algorithms. Experiments in perpendicular and parallel parking scenarios show that N3P-enhanced Hybrid A* speeds up planning by more than 80\%. 
It also outperforms RL baselines in success rate and trajectory quality, producing shorter trajectories with fewer gear changes, while achieving comparable or lower planning time in most cases.
\end{abstract}
%%%%%%%%%%%%%%%%%%%%%%%%%%%%%%%%%%%%%%%%%%%%%%%%%%%%%%%%%%%%%%%%%%%%%%%%%%%%%%%%
\section{INTRODUCTION}
Autonomous parking is an important Driver Assistance function in modern cars. For autonomous parking to perform well in real life, we need algorithms to generate kinematically feasible, collision-free paths. Ideally, such an algorithm should provide low-latency planning suitable for practical parking scenarios with tight spatial constraints.

Graph-search and optimization-based algorithms are among the most widely used approaches for autonomous parking path planning. Classical graph-search methods such as A* and Rapidly-exploring Random Trees (RRT) guarantee theoretical completeness, 
%i.e., they will find a solution if one exists,
but cannot ensure kinematic feasibility of the resulting path \cite{cheng2014improved, han2011unified, a_star,lavalle1998rapidly}. Hybrid A* addresses this by incorporating motion primitives that satisfy nonholonomic constraints, often augmented with analytic expansions using Reeds-Shepp curves to improve accuracy and reduce node expansion \cite{dolgov2008practical,sussmann1991shortest, fast_rs}.
% , enabling it to handle non-holonomic constraints and avoid obstacles. 
Nevertheless, in tightly constrained parking scenarios, Hybrid A* and its sped-up variants  
% e.g., SHA* \cite{sha_satr}, MHHA* \cite{huang2022search}, Informed Hybrid A* \cite{informed_ha}, 
still require extensive node expansions and rarely achieve sub-second runtimes \cite{sha_satr, huang2022search, informed_ha}. 
Optimization-based methods generate smooth, dynamically feasible, and cost-efficient trajectories. However, they are susceptible to convergence to local minima \cite{zhang2020optimization, mppi, mpcpathplanning} and often require several seconds—or even minutes—to converge, even when a globally optimal solution exists. To mitigate local minima issues inherent in nonlinear path optimization, popular approaches initialize optimization \cite{zhang2020optimization, zhang2025automatic, LI2024104816, z_han} or construct convex corridors from nonconvex obstacle constraints \cite{optimization2022guided} using coarse trajectories generated by Hybrid A*. Consequently, graph-search computation remains a major contributor to the overall runtime bottleneck in such optimization frameworks.

More recently, learning-based methods have been proposed, leveraging deep neural networks \cite{chai2020design, chai2022deep, kim2023neural} and reinforcement learning (RL) \cite{shen2023multi, chai2022design, jiang2025hope} to handle highly constrained parking. Although these methods are fast at inference, %particularly when deployed on specialized hardware such as GPUs. However, 
their performance depends strongly on the quality, diversity, and coverage of training data, and distributional shifts can significantly degrade reliability. In addition, learned policies typically do not enforce explicit recursive feasibility, which can lead to failure in reaching the goal and aggressive near-collision maneuvers, reducing passenger comfort. These limitations motivate exploring alternative approaches that combine human-inspired strategies with structured planners. 
% In particular, by incorporating naturalistic driving behaviors into a graph-search or optimization-based framework, it may be possible to achieve reliable convergence and high-quality trajectories under reasonable runtime, without relying solely on learned policies.

\begin{figure}
    \centering
    \includegraphics[width=0.49\linewidth, trim=0pt 30pt 0pt 30pt, clip]{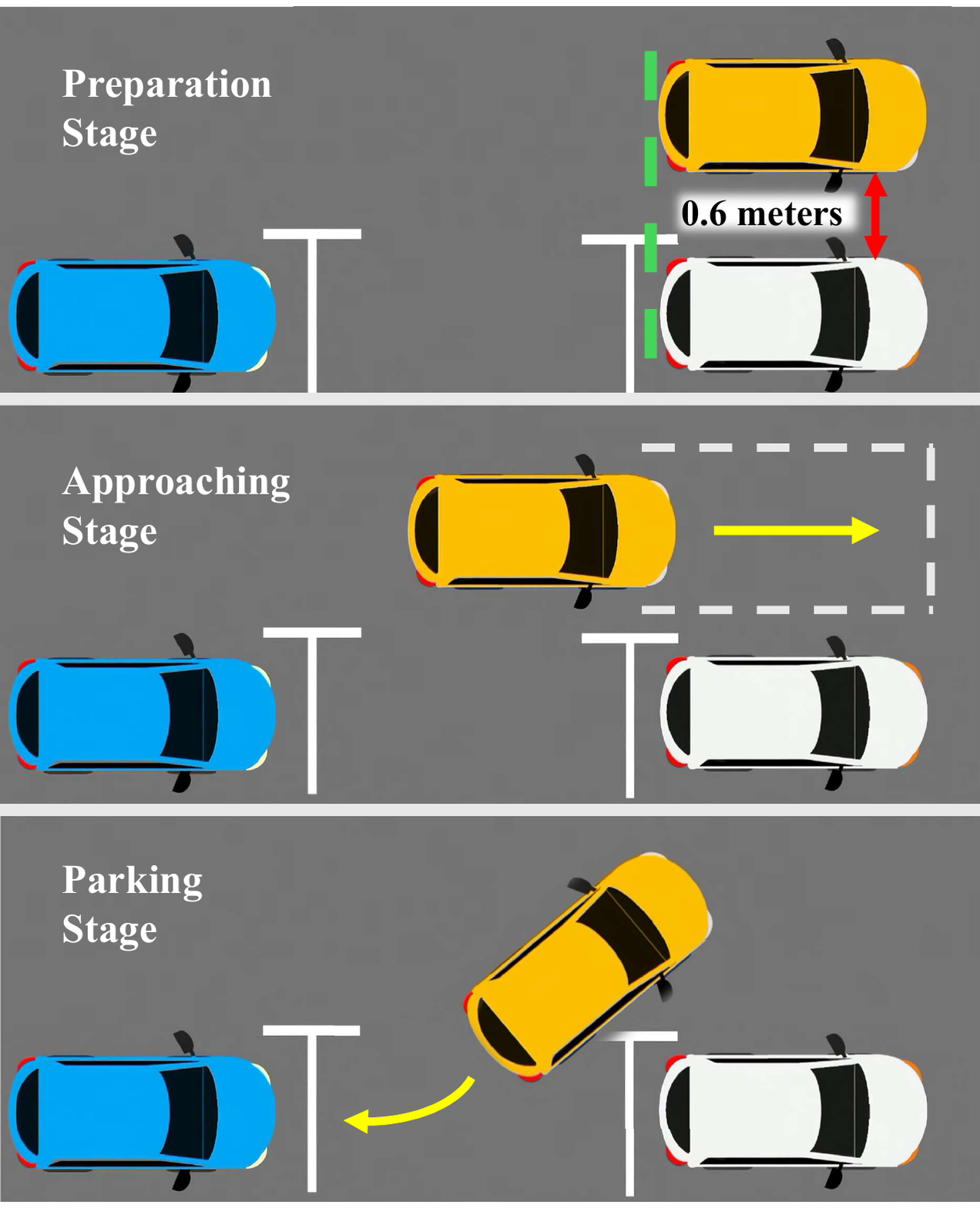} 
    \caption{During parking, human drivers often follow a preparation phase to select an intermediate pose, an approaching phase to reach it, and a parking phase to complete the maneuver. N3P adopts this strategy, using an offline-learned model to select preparatory poses and accelerate planning.}
    \vspace{-15pt}
    \label{fig:n3p}
\end{figure}
Unlike traditional autonomous path planners, which exhaustively explore all possible trajectories from the vehicle’s current pose to the parking spot to find an optimal path, human drivers often follow a more efficient three-stage strategy as illustrated in \autoref{fig:n3p}. In the first \textit{preparation stage}, drivers identify an intermediate pose near the target parking spot that facilitates an easier final maneuver. In the second \textit{approaching stage}, they maneuver the vehicle into this preparatory pose. Finally, in the third \textit{parking phase}, they execute a smooth, continuous motion from the preparatory pose into the parking spot. Motivated by this behavior, we propose N3P, a naturalistic three-stage parking scheme that uses an offline-learned model to select preparatory poses during planning, accelerating the overall process. 

% \textbf{Literature Review:} 
Dividing complex path planning problems into smaller subproblems is not new in parking control architecture design. For example, \cite{BOS20234877} decomposes MPC-based planning into sequential traversal of three predefined rectangular corridors, while \cite{zhang2023hierarchical} analytically computes a preparatory pose for forward and parallel parking based on vehicle geometry and steering limits. However, these approaches do not scale well to complex valet parking scenarios. The corridor structure in \cite{BOS20234877} is validated only on one perpendicular and one parallel parking case with manually designed corridors, and it is unclear how such corridors should be adapted to varying initial poses and environments. Although \cite{zhang2023hierarchical} improves generalization, it is still restricted by the limited representational capacity of analytical methods in environments with spacious, obstacle-free lanes. This limitation prevents its application in real-world valet parking settings with irregular layouts and surrounding vehicles.

The contributions of the paper are summarized as follows: 
\begin{itemize}
% \item \yifan{We propose a three-stage parking scheme that imitates naturalistic human parking strategies: (i) identify a preparatory pose given the environment $\mathcal{E}$, (ii) approach the pose, and (iii) execute the final steering maneuver into the target.
\item  A naturalistic three-stage parking framework that reduces computation complexity of parking path planning via a learned preparatory pose selector.
\item A scalable offline-to-online abstraction pipeline that enables a learned preparatory pose selector, trained on simplified environments, to generalize to realistic cluttered parking scenes.
\end{itemize}
% \item To ensure scalability and generalization to realistic parking scenarios, we design a novel offline training and online adaptation pipeline that enables the preparatory-pose selector, trained on simplified environments, to perform robustly in diverse, real-world layouts.
% \item Within this scheme, we leverage the starting poses of Reeds–Shepp path segments on drivable trajectories, generated by Reeds–Shepp–augmented Hybrid A*, as candidate preparatory poses, providing a systematic method to generate large-scale training data and to integrate classical path planning with the three-stage framework.
% \item Learning-based preparatory-pose selection kernels that predict preparatory poses from environment features using a Multilayer Perceptron (MLP) and K-Nearest Neighbors (KNN) algorithms are developed.
% \item We validate the three-stage scheme by integrating it with Hybrid A*. The N3P-enhanced Hybrid A*'s performance is compared against various Hybrid A* variants and HOPE \cite{jiang2025hope}, a transformer-based RL agent enhanced with Reeds-Shepp paths, using randomly generated parking scenarios that emulate real-world challenges such as tight maneuvering spaces and dead ends.
To evaluate N3P in a representative and widely adopted setting, we integrate it within Hybrid A*, which is commonly used both as a standalone planner and as an initialization for optimization-based control. The resulting N3P-enhanced Hybrid A* is compared against existing Hybrid A* variants and HOPE \cite{jiang2025hope}, a transformer-based RL agent, using randomly generated parking scenarios that emulate real-world challenges such as tight maneuvering spaces and dead ends. 

\section{Problem formulation}
\subsection{Environment Construction}
\begin{figure}[htbp]
    \centering
    % First subfigure
    \begin{subfigure}[b]{0.62\linewidth}
        \centering
        \includegraphics[width=\linewidth, trim=5pt 43pt 78pt 50pt, clip]{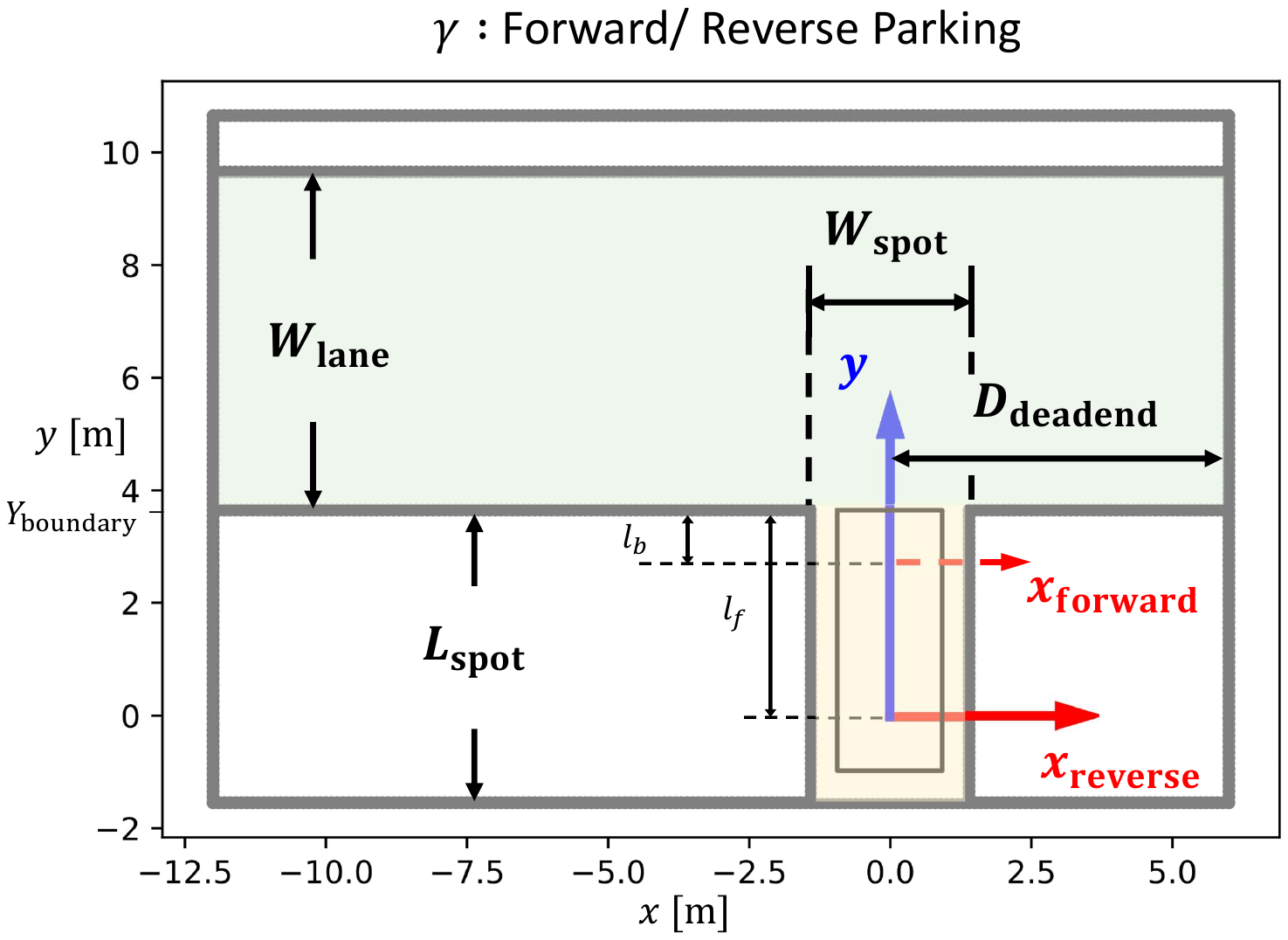} 
        % \caption{\label{fig:reverse env}}
    \end{subfigure}
    % Second subfigure
    \begin{subfigure}[b]{0.62\linewidth}
        \centering
        \includegraphics[width=\linewidth, trim=5pt 80pt 10pt 40pt, clip]{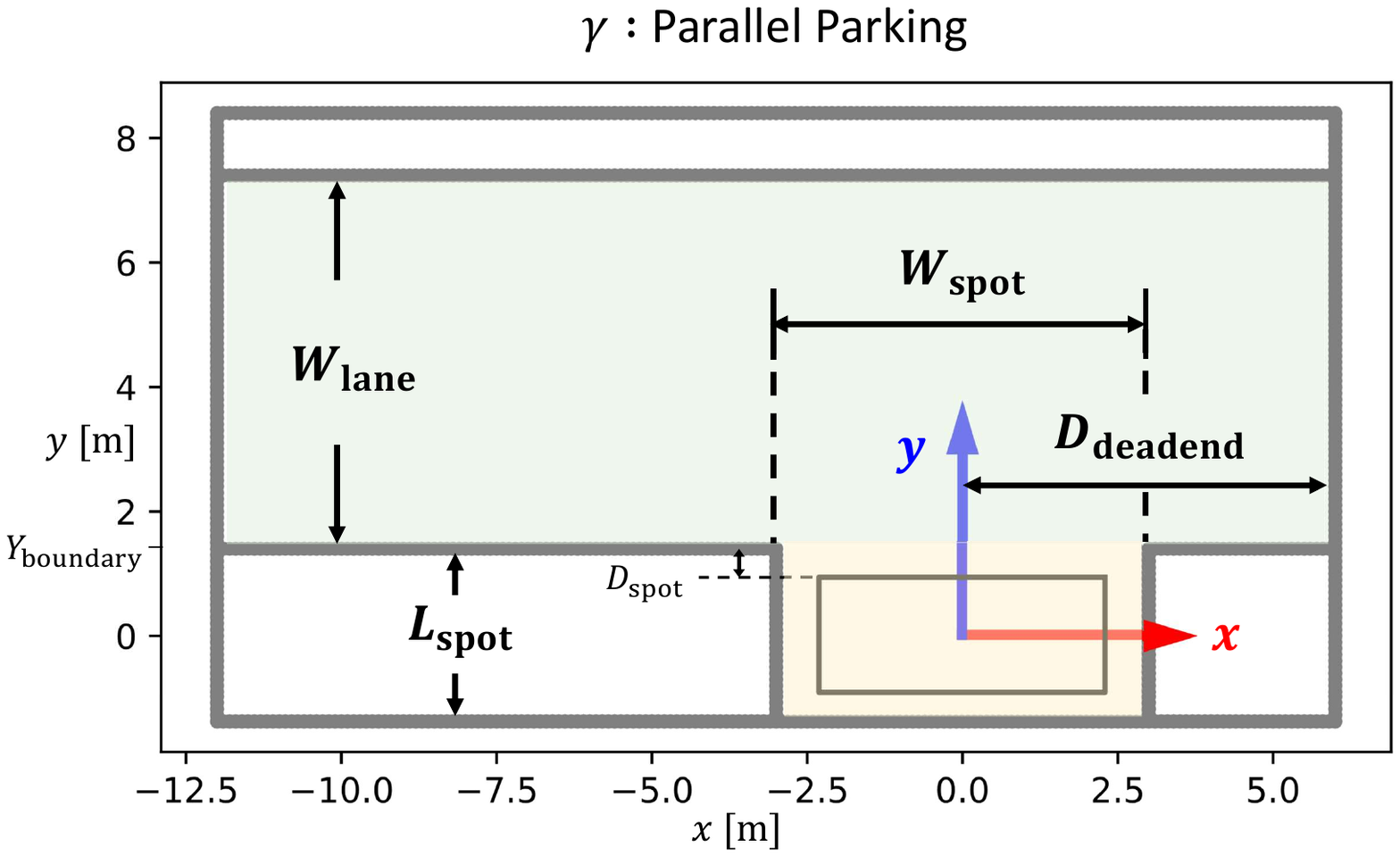} 
        % \caption{\label{fig:parallel env}}
    \end{subfigure}
    \caption{Examples of parking-spot-based frames $\hat{p}$ for forward, reverse, and parallel parking. Drive lanes are shown in green and parking spots in yellow.}
    \vspace{-10pt}
    \label{fig:env}
\end{figure}

Without loss of generality, we formulate the problem in a parking-spot-based frame $\hat{p}$, where the parking spot center is set as the origin, as illustrated in Fig.~\ref{fig:env}. The coordinate system is defined such that the positive $x$-axis aligns with the drive lane and points toward the vehicle’s goal heading for parallel parking, while the positive $y$-axis points from the parking spot toward the drive lane. We denote the parking goal poses as $\boldsymbol{g}^{\hat{p}}$.
% \vspace{-10pt}
% \begin{equation}
% Y_\text{boundary}(\gamma) = 
% \begin{cases}
%     l_f  \quad \textbf{if} \quad \gamma=\text{\say{forward}}\\
%     l_b  \quad \textbf{if} \quad \gamma=\text{\say{reverse}} \\
%     D_\text{spot}+w/2 \quad \textbf{if} \quad \gamma=\text{\say{parallel}}
% \end{cases}
% \label{eq:y boundary}
% \end{equation}

% \begin{equation}
% \boldsymbol{g}^{\hat{p}}(\gamma) = 
% \begin{cases}
%     [0, 0, -\pi/2]^\top \quad \quad \textbf{if} \quad \gamma=\text{\say{forward}}\\
%     [0, 0, \pi/2]^\top \quad \quad \textbf{if} \quad \gamma=\text{\say{reverse}}\\
%     [0,l_c, 0]^\top \quad \textbf{if} \quad \gamma=\text{\say{parallel}}
% \end{cases}
% \label{eq:goal}
% \end{equation}

\subsection{Vehicle Model}
Let $(x,y, \theta)$ be the autonomous vehicle state, where $x, y$ are the 2D positions in meters and $\theta$ is the angle between the vehicle heading and the positive $x$-axis in radians.  We employ the kinematic bicycle model for vehicle dynamics, which is well-suited for vehicles at low speeds \cite{rajamani2006vehicle} and has been widely used in similar autonomous parking applications \cite{jiang2025hope, dolgov2008practical, guided_ha, informed_ha, sha_satr}.
\begin{equation}
\boldsymbol{\dot{x}} = 
\begin{bmatrix}
    \dot{x}\\ \dot{y}\\ \dot{\theta}
\end{bmatrix} =
\begin{bmatrix}
    v\cos{\theta}\\v\sin{\theta}\\ \frac{v}{l}\tan{\delta} 
\end{bmatrix}
\label{eq: vehicle model}
\end{equation}
\yifan{Here $l$ is the wheelbase of the vehicle, approximated as a rectangle. $u=[v, \delta]^\top$ is the control input, where $v$ is the longitudinal velocity and $\delta$ is the steering angle of the front wheel. The control inputs are subject to box constraints: $v\in [v_\text{min}, v_\text{max}] $ }and $\delta \in [\delta_\text{min}, \delta_\text{max}]$.
% To simplify notations, we denote the distance from the rear wheel axis to the front end of the vehicle as $l_f$, that to the vehicle's back end as $l_b$, and the width of the vehicle as $w$. Then the perpendicular distance from the vehicle center to the rear wheel axis of the vehicle $l_c$ and the vehicle length $l_\text{car}$ can be computed as follows.
% \begin{gather}
% l_c=(l_f-l_b)/2\\
% l_\text{car} = l_f+l_b
% \label{eq:vehicle dim}
% \end{gather}
\subsection{Reeds-Shepp Path}
% In \cite{reeds1990optimal}, James Reeds and Lawrence Shepp proved that in obstacle-free environment, the shortest path between two planar vehicle configurations $(x, y, \theta)$ for a car-like vehicle -- with bounded curvature $\kappa = \tan(\delta_{\max})/l$ and bidirectional motion -- most consists of a finite sequence of motion primitives composed of straight segments ($S$) and maximum-curvature circular arcs ($L$ for left turns and $R$ for right turns). Such path is referred to a Reeds-Sheep (RS) path. Reeds-Shepp path can be extended to environemnts with obstacles by.... However, in none-obstacle-free environment, finding a collision-free reed-shepp path connecting two vehicle configurartions is not guarantueed.

% Due to its analytical tractability and ability to model bidirectional motion, RS paths can naturally represent multi-point parking behaviors such as parallel parking, perpendicular parking, and garage parking. 
Reeds and Shepp \cite{reeds1990optimal} proved that, in an obstacle-free environment, the shortest path between planar configurations $(x, y, \theta)$ for a bidirectional vehicle with bounded curvature $\kappa = \tan(\delta_{\max})/l$ consists of a finite sequence of straight segments ($S$) and maximum-curvature arcs ($L$ for left turns and $R$ for right turns). Such trajectories are commonly referred to as Reeds-Shepp (RS) paths. In environments with obstacles, collision-free RS paths are obtained by generating candidate RS curves between the initial and goal configurations and performing collision checking on each trajectory. Due to its analytical tractability and ability to model bidirectional motion, RS paths are widely used as a local connection strategy in autonomous parking algorithms.

% For any pair of configurations, the shortest feasible path belongs to a finite family of candidate primitive combinations and can be obtained analytically by evaluating this family and selecting the minimum-length solution. Due to its analytical tractability and ability to model bidirectional motion, the Reeds-Shepp path is widely used as a local connection strategy in autonomous parking algorithms.

\subsection{Path Planning Problem Definition}
We assume the autonomous vehicle has access to detectable obstacle point clouds along the path from its initial position to the parking target. The scope of this work focuses on \textbf{perpendicular} (forward and reverse) and \textbf{parallel} parking scenarios, but this approach could be extended to include additional parking types such as angled parking.

% \yifan{\textbf{Cost Functions:} The quality of each candidate path is evaluated by the total cost $C$, defined as the sum of motion costs $c(u_t, u_{t-1})$ that penalize gear changes, steering magnitude, steering changes, and distance traveled:
% \begin{align}
%     c(u_t, u_{t-1}) = &\lambda_\text{dc}\mathbf{1}[{\operatorname{sign}(v_t) \neq \operatorname{sign}(v_{t-1})}]+\lambda_\text{arc}(v)l_\text{arc}\\
% &\nonumber +\lambda_\text{steer}|\delta_t|+\lambda_{\Delta}|\delta_t-\delta_{t-1}|,
% \end{align}
% where $l_\text{arc}$ is the arc length of the motion primitive, $\lambda_\text{dc}$, $\lambda_\text{steer}$, and $\lambda_{\Delta}$ are the cost coefficients for gear-switching (direction-change), steering magnitude, and steering change. Lastly, 
% \begin{equation}
%     \lambda_\text{arc}(v) =
% \begin{cases}
% \lambda_\text{f}, & v \ge 0, \\
% \lambda_\text{b}, & v < 0,
% \end{cases}
% \end{equation}
% is direction-dependent driving cost, where $\lambda_\text{f}$ and $\lambda_\text{b}$ denote the per-unit-length forward and backward driving costs.}

\textbf{Problem Statement}: 
Given the initial state $\boldsymbol{x}_0^{\hat{p}}$ of the ego vehicle, and obstacle point clouds in the parking environment, the goal is to find a sequence of control inputs $u$ that satisfy the kinematic model in \autoref{eq: vehicle model} and the actuation limits, steering the vehicle to the desired parking pose $\boldsymbol{g}^{\hat{p}}\in \mathbb{R}^3$ while avoiding obstacle collisions and yielding a path of short trajectory length, few gear changes and few steering angle changes.

\section{Methodology Overview}
% Planning a preparatory pose that enables a smooth turn into a parking spot lets drivers simplify complex parking into intuitive steps. Inspired by this natural human parking behavior, the N3P framework we proposed consists of three stages:
Inspired by how human drivers use a preparatory pose to simplify parking into intuitive steps, the N3P framework we propose similarly consists of three stages:
\begin{itemize}
    \item Preparation stage: Given an arbitrary parking environment, abstract the region surrounding the goal parking pose into a local configuration $\mathcal{E}$ and determine a feasible preparatory pose $\boldsymbol{g}_\text{pre}^{\hat{p}} \in \mathbb{R}^3$ based on this abstraction.
    \item Approaching stage: Solve for a path from the initial vehicle state $\boldsymbol{x}_0^{\hat{p}}$ to the preparatory pose $\boldsymbol{g}_\text{pre}^{\hat{p}}$ using a graph-search or optimization-based parking planner.
    \item \yifan{Parking stage}: Compute an optimal parking maneuver from the preparatory pose $\boldsymbol{g}_\text{pre}^{\hat{p}}$ to the parking goal $\boldsymbol{g}^{\hat{p}}$ using an analytical Reeds–Shepp path.
\end{itemize}
Selecting preparatory poses faces three key challenges. First, parking environments vary in spot dimensions, infrastructure geometry, and surrounding vehicles, creating effectively unlimited scenarios. Abstracting these diverse conditions into a finite set that can be handled with limited offline data and training is nontrivial. Second, poorly selected intermediate poses can degrade path quality, resulting in longer trajectories or more direction changes. 
Third, ensuring feasibility is challenging: although planning from $\boldsymbol{x}_0^{\hat{p}}$ to $\boldsymbol{g}_\text{pre}^{\hat{p}}$ is typically more efficient than planning directly to $\boldsymbol{g}^{\hat{p}}$, the subsequent connection from $\boldsymbol{g}_\text{pre}^{\hat{p}}$ to $\boldsymbol{g}^{\hat{p}}$ may still be computationally expensive or even infeasible under kinematic and environmental constraints.
In the following sections, we describe how the N3P framework addresses these challenges. \autoref{sec:simplify env} presents how diverse parking environments are abstracted into a finite set of representative configurations. \autoref{sec:pose selector} describes how offline data generation and model training produce preparatory poses that allow smooth and easily executed maneuvers to the parking goal. Finally, \autoref{sec:pha} explains the online deployment of the learned preparatory pose selector for accelerated path planning in realistic parking scenarios.
\begin{figure}
    \centering
    \includegraphics[width=1\linewidth, trim=0 180 0 190, clip]{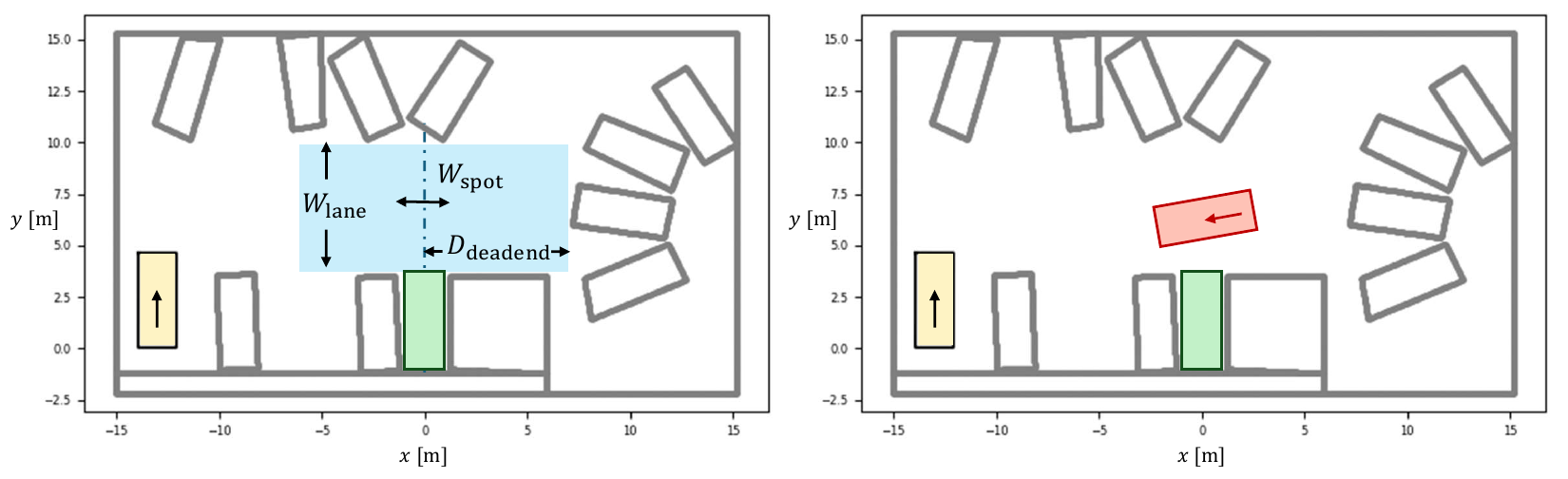}
    % \vspace{-10pt}
    \caption{Left: extraction of abstracted parameters from an arbitrary environment under valid parking conditions. Right: an invalid scenario where another vehicle blocks the rectangular access region of the target parking space.}
    \label{fig:abstraction}
    \vspace{-10pt}
\end{figure}
\section{Environment Abstraction}
\label{sec:simplify env}
Obstacle distributions near a parking goal vary widely, making it impractical to explicitly model all possible configurations or train a preparatory pose selector that accounts for every geometric detail. Rather than attempting exhaustive coverage of obstacle layouts, we instead identify the geometric conditions that fundamentally determine the type of maneuver required. As shown in \autoref{fig:abstraction}, we observe that a parking goal is feasible only if a sufficiently large obstacle-free rectangular region exists on the drive lane adjacent to the parking spot. The size and relative position of this free-space rectangle directly determines the maneuver strategy, for example, whether ample approach space is available or whether constrained, multi-switch maneuvers are necessary. Therefore, instead of representing full obstacle configurations, we abstract each environment by the geometry of its largest feasible free-space rectangle. Under this abstraction, two environments are considered equivalent for preparatory pose selection if they share the same free-space rectangle dimensions and relative positioning under the same parking type. This reduction enables us to exhaustively cover all geometrically distinct maneuver conditions using only four parameters: lane width $W_\text{lane}$, spot width $W_\text{spot}$, dead-end depth $D_\text{deadend}$, and parking type $\gamma$ (forward, reverse, or parallel), as in \autoref{fig:env}. 
\begin{equation}
    \mathcal{E} = \{W_\text{lane}, W_\text{spot}, D_\text{deadend}, \gamma\}
    \label{eq:env def}
\end{equation}
The parking type $\gamma$ is assumed known from the parking-spot geometry prior to local planning. When the parking lane does not have a nearby dead end, $D_\text{deadend}$ is the distance from the center of the vehicle parking spot to the boundary of our local parking environment. The parking spot length $L_\text{spot}$ is not used to describe the parking infrastructure because autonomous vehicles do not need this information to determine their parking poses. In the perpendicular parking scenarios, once the vehicle finds a path to the poses shown in \autoref{fig:env}, it can easily adjust itself to the desired depth inside the parking spot by simply moving forward or backward in a straight line. In parallel parking scenarios, any final parking pose is considered appropriate as long as it leaves a sufficient depth $D_\text{spot}$ from the boundary $Y_\text{boundary}$ between the parking spot and the lane.
% We further assume that given arbitrary local parking environment where a kinematically feasible and collision-free path exist connecting the vehicle initial position and the target, the users have a module that can in real-time find the simplified version of the environment described using four environment abstraction parameters, as example of which is provided in Fig.~\ref{fig:simplify env}.
% The abstraction parameters remain fixed throughout the navigation process, while the point clouds may vary due to the presence of dynamic obstacles such as pedestrians or other vehicles. 

\section{Preparatory Pose Selector}
\label{sec:pose selector}
A good preparatory pose $\boldsymbol{g}_\text{pre}^{\hat{p}}$ in parking navigation should satisfy two requirements: (1) there must exist a kinematically feasible and collision-free trajectory from the vehicle’s initial state $\boldsymbol{x}_0^{\hat{p}}$ to $\boldsymbol{g}_\text{pre}^{\hat{p}}$, and (2) it should be straightforward to generate a smooth connecting path from $\boldsymbol{g}_\text{pre}^{\hat{p}}$ to the final parking pose $\boldsymbol{g}^{\hat{p}}$.  To satisfy the second requirement, we consider RS paths, which naturally capture the multi-segment forward–reverse maneuvers required in parking and enable real-time closed-form evaluation of the connecting trajectory. The remaining challenge is to identify start poses that admit collision-free RS connections to $\boldsymbol{g}^{\hat{p}}$ while also being reachable from the initial state $\boldsymbol{x}_0^{\hat{p}}$. A straightforward approach is to uniformly sample configurations around the parking spot and evaluate their connections. However, since RS paths are defined over a continuous configuration space, valid connecting configurations are typically confined to a narrow subset of states, making them difficult to reliably discover through discrete sampling, even at fine resolutions.

Here, we novelly select the \textbf{starting pose of the RS curve} $\boldsymbol{x}_\text{rs}^{\hat{p}}$ generated by the RS-augmented Hybrid A* as exemplary preparatory poses for parking applications \cite{dolgov2008practical}. 
Since $\boldsymbol{x}_\text{rs}^{\hat{p}}$ is a point on the drivable Hybrid A* path computed, it trivially ensures a kinematically feasible and collision-free path to $\boldsymbol{x}_0^{\hat{p}}$, as well as to $\boldsymbol{g}^{\hat{p}}$. 
% \yifan{Moreover, using the start poses of Reeds-Shepp paths also satisfies the second requirement, as given two poses that are guaranteed to have a collision-free Reeds-Shepp path connecting one another, computing that path analytically takes less than a millisecond.}

% Given an arbitrary parking environment, human drivers estimate suitable preparatory poses using their past parking experience. Imitating human drivers' preparatory pose estimation behavior, we collect $\tilde{\boldsymbol{x}}_\text{rs}^{\hat{p}}$ by solving Reeds-Shepp-enhanced Hybrid A* offline and use the data collected to select preparatory poses online through learning-based approaches. 

\subsection{Offline Data Collection}
\label{sec:offline data}
% To make our method generalizable to different parking scenarios and parking environments, 
We categorize parking scenarios by environment parameters $\mathcal{E}$ and construct a collective set $\mathbb{E}$: 
\begin{equation}
\begin{aligned}
\mathbb{E} = \{\mathcal{E}| \forall W_\text{lane} \in \mathcal{W}_\text{lane},W_\text{spot} \in \mathcal{W}_\text{spot}, \\D_\text{deadend} \in \mathcal{D}_\text{deadend}, \gamma \in \Gamma_\text{type}\},
\end{aligned}
\end{equation}
where $\mathcal{E}$ is single parking scenario defined as in \autoref{eq:env def}, \yifan{and $\mathcal{W}_\text{lane}$, $\mathcal{W}_\text{spot}$, and $\mathcal{D}_\text{deadend}$ denote the discretized sets of drive-lane width, parking-spot width, and dead-end length, respectively. Each set is uniformly sampled within a predefined range with a fixed resolution, reflecting typical dimensions in real-world parking structure designs.}
\begin{gather}
\nonumber \mathcal{W}_\text{lane}=\left\{ w^{(l)} \mid i \in \mathbb{N},\,  w^{(l)} =  w^{(l)}_\text{min} + i\Delta w^{(l)} \leq  w^{(l)}_\text{max} \right\},\\
\nonumber \mathcal{W}_\text{spot}=\left\{  w^{(s)} \mid i \in \mathbb{N},\,  w^{(s)} =  w^{(s)}_\text{min} + i\Delta w^{(s)} \leq  w^{(s)}_\text{max} \right\},\\ 
\nonumber\mathcal{D}_\text{deadend} = \left\{ d \mid i \in \mathbb{N},\, d = d_\text{min} + i\Delta d\leq d_\text{max} \right\},\\
\gamma \in \Gamma_\text{type} = \{\text{\say{forward}, \say{reverse}, \say{parallel}}\}.
\label{eq:E set def}
\end{gather}
Here, $(w^{(s)}_\text{min}, w^{(s)}_\text{max}),(w^{(l)}_\text{min}, w^{(l)}_\text{max}),(d_\text{min}, d_\text{max})$ are respectively the range of $W_\text{lane}$, $W_\text{spot}$, $D_\text{deadend}$ to be considered and $\Delta w^{(s)},\Delta w^{(l)},\Delta d$ are corresponding resolutions. The environment simplification in \autoref{sec:simplify env}  enables representing obstacle conditions with only three parameters—$W_\text{lane}$, $W_\text{spot}$, and $D_\text{deadend}$. By uniformly sampling within defined parameter ranges, the data are guaranteed to represent all distinctive environments of interest with controllable granularity. Such completeness cannot be ensured by high-dimensional environment representations based on obstacle configurations or LiDAR distance vectors. Although the parameter ranges must be specified by the user, the calibration burden is minimal, as they can be readily determined from local regulations on lane widths and parking spot dimensions.

\begin{figure}
    \centering
    \includegraphics[width=0.61\linewidth, trim=0pt 16pt 0pt 14pt, clip]{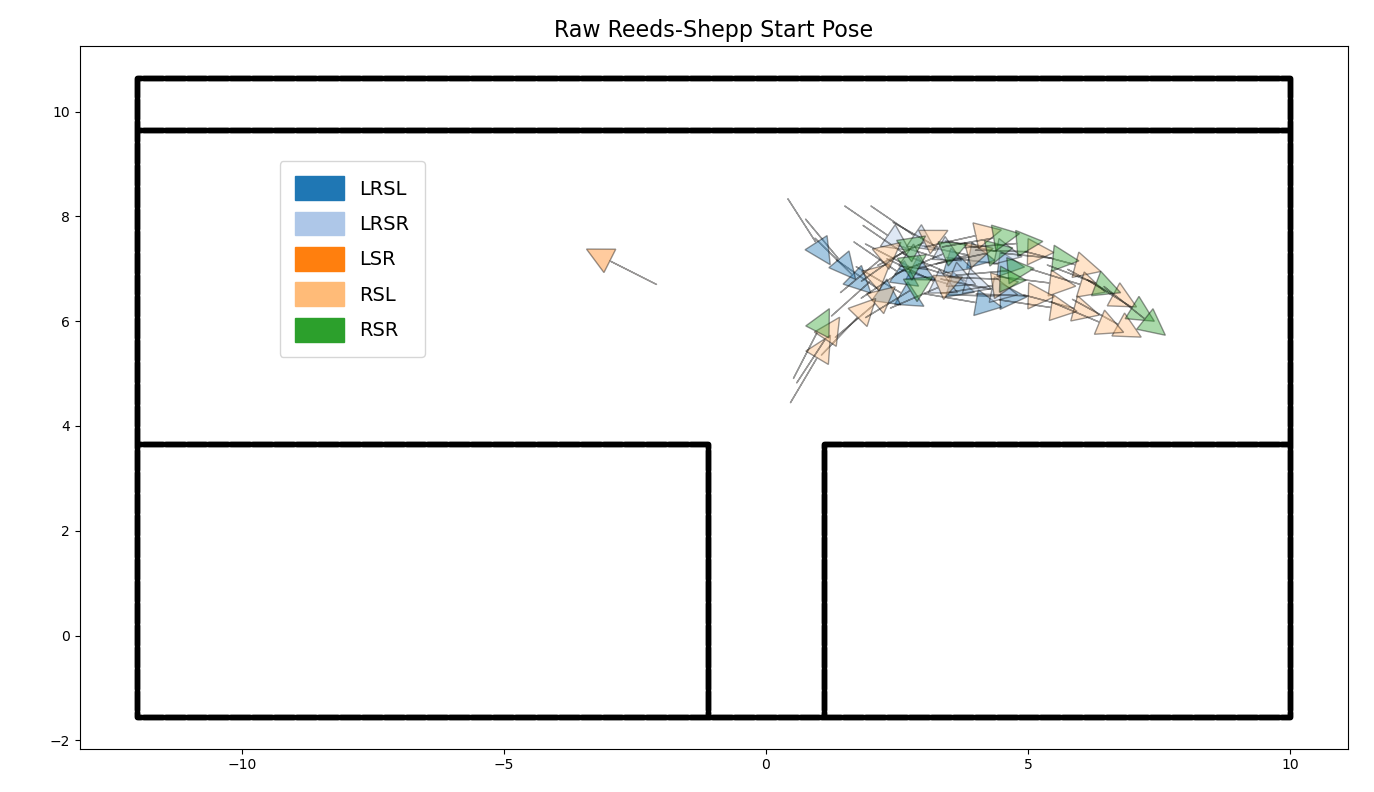}
    % \vspace{-3pt}
    \caption{Reeds–Shepp start poses for the reverse parking environment $\mathcal{E}$, colored by curve type to the goal: “S” for straight, “R”/“L” for right/left arcs.}
    \label{fig:rs poses}
    \vspace{-10pt}
\end{figure}

For each parking scenario in $\mathbb{E}$, we repeatedly solve the Hybrid A* augmented with RS paths offline using initial states $\tilde{\boldsymbol{x}}^{\hat{p}}_0$ uniformly sampled in the free space of the abstracted environments. The starting poses $\tilde{\boldsymbol{x}}^{\hat{p}}_\text{rs}$ of the RS segments in the resulting drivable trajectories are stored in the set $\tilde{\mathcal{X}}^{\hat{p}}_\text{rs}[\mathcal{E}] \in \mathbb{R}^{N_{\mathcal{E}}\times 3}$, with their corresponding initial states $\tilde{\boldsymbol{x}}^{\hat{p}}_0$ stored in $ \tilde{\mathcal{X}}^{\hat{p}}_0[\mathcal{E}] \in \mathbb{R}^{N_{\mathcal{E}}\times 3}$, where $N_{\mathcal{E}}$ is the number of data samples collected for parking environment $\mathcal{E}$. 
% The initial poses $\tilde{\mathcal{X}}^{\hat{p}}_0$ are uniformly sampled in the free space of the abstracted environments. 
% as in \autoref{eq:offline sample}, where $N \in \mathbb{R}^+$ is a user-defined positive number, and $Y_\text{boundary}$ can be computed using \autoref{eq:y boundary}.
% \begin{equation}
% \begin{aligned}
%     \tilde{\boldsymbol{x}}^{\hat{p}}_0=[\tilde{x}^{\hat{p}}_0, \; \tilde{y}^{\hat{p}}_0, \; \tilde{\theta}^{\hat{p}}_0] \quad \quad -\pi/6 \leq \tilde{\theta}^{\hat{p}}_0 \leq \pi/6\\
%     -N\times l_\text{car}\leq \tilde{x}^{\hat{p}}_0\leq \min(N\times l_\text{car}, D_\text{deadend})\\
%     Y_\text{boundary}(\gamma) \leq \tilde{y}_0 \leq Y_\text{boundary}(\gamma)+W_\text{lane}
%     \label{eq:offline sample}
% \end{aligned}
% \end{equation}
% The initial heading angles of the vehicle are set to be in $[-\pi/6, \pi/6]$ radians; vehicles are typically aligned with the parking lane when approaching to a parking spot in the same corridor. 
Examples of collected $\tilde{\mathcal{X}}^{\hat{p}}_\text{rs}[\mathcal{E}]$ can be found in \autoref{fig:rs poses}. 

\subsection{Learning-Based Models}
In the learning-based model, we trained models using collected RS starting poses $\tilde{\mathcal{X}}^{\hat{p}}_\text{rs}$, vehicle initial states $\tilde{\mathcal{X}}^{\hat{p}}_0$ and the environment parameters $\mathbb{E}$ corresponded to them. Given the vehicle's initial states $\boldsymbol{x}_0^{\hat{p}}$ and the parking environment $\mathcal{E}$, as inputs, the goal of our learned preparatory pose selector $f_{\gamma}(\cdot)$ can be formally summarized as follows:
\begin{gather}
    \boldsymbol{g}^{\hat{p}}_\text{pre}(\boldsymbol{x}_0^{\hat{p}},\mathcal{E}) = f_{\gamma}(\mathbf{x};\mathbf{w}), \\
    \mathbf{x} = [x^{\hat{p}}_0,y^{\hat{p}}_0,\theta^{\hat{p}}_0,W_\text{lane}, W_\text{spot}, D_\text{deadend}]^\top,
    \label{eq:input}
\end{gather}
where $\mathbf{x}\in \mathbb{R}^6$ is the input, $\mathbf{w}$ is the parameter vector, and $\boldsymbol{g}_\text{pre}^{\hat{p}}$ is the preparatory pose output of the model. The model $f_{\gamma}(\mathbf{x}; \mathbf{w})$ is trained using input features $\mathbf{X}^{(\gamma)}_\text{train}$ and training targets $\mathbf{Y}^{(\gamma)}_\text{train}$, conditioned on parking type $\gamma \in \Gamma_\text{type}$. The training data are concatenated using $ \tilde{\mathcal{X}}^{\hat{p}}_\text{rs},  \tilde{\mathcal{X}}^{\hat{p}}_0$ collected in parking environments from set $\mathbb{E}_{\gamma}\subset \mathbb{E}$, where all the environments share the same parking type $\gamma$. \yifan{Denote $\mathcal{E}_i$ as the $i$-th element in $\mathbb{E}_{\gamma}$, $N_{\mathcal{E}_i}$ as the size of $\tilde{\mathcal{X}}^{\hat{p}}_0[\mathcal{E}_i]$ (or equivalently, the size of $\tilde{\mathcal{X}}^{\hat{p}}_\text{rs}[\mathcal{E}_i]$), and the number of environment configuratins in $\mathbb{E}_{\gamma}$ as $|\mathbb{E}_{\gamma}|$. Let $\mathcal{E'}_i=[W_\text{lane}^{(i)}, W_\text{spot}^{(i)}, D_\text{deadend}^{(i)}]$ be the vector of $\mathcal{E}_i$ excluding parking type $\gamma$, and $\mathbf{1}_{N_{\mathcal{E}_i}}$ be the $N_{\mathcal{E}_i}$-dimensional vector of ones, we get
\begin{equation}
\nonumber \mathbf{X}_\text{train}^{(\gamma)} = 
\begin{bmatrix}
    \tilde{\mathcal{X}}^{\hat{p}}_0[\mathcal{E}_1] & \mathbf{1}_{N_{\mathcal{E}_{\scriptscriptstyle 1}}}\mathcal{E'}_1^\top\\
    \tilde{\mathcal{X}}^{\hat{p}}_0[\mathcal{E}_2] & \mathbf{1}_{N_{\mathcal{E}_{\scriptscriptstyle 2}}}\mathcal{E'}_2^\top\\
    \vdots & \vdots\\
    \tilde{\mathcal{X}}^{\hat{p}}_0[\mathcal{E_{|\mathbb{E_{\gamma}}|}}]& 
    \mathbf{1}_{N_{\mathcal{E}_{|\mathbb{E}_{\gamma}|}}}\mathcal{E'}_{|\mathbb{E}_{\gamma}|}^\top
\end{bmatrix}, \;\;\mathbf{Y}_\text{train}^{(\gamma)} = 
\begin{bmatrix}
  \tilde{\mathcal{X}}^{\hat{p}}_\text{rs}[\mathcal{E}_1]\\
  \tilde{\mathcal{X}}^{\hat{p}}_\text{rs}[\mathcal{E}_2]\\
  \vdots\\
  \tilde{\mathcal{X}}^{\hat{p}}_\text{rs}[\mathcal{E_{|\mathbb{E_{\gamma}}|}}]
\end{bmatrix} .
\end{equation}}

% \begin{equation}
% \mathbf{X}_\text{train} = 
% \begin{bmatrix}
%     \tilde{\mathcal{X}}^{\hat{p}}_0[\mathcal{E}_1] & W_\text{lane}^{(\mathcal{E}_1)} & W_\text{spot}^{(\mathcal{E}_1)} & D_\text{deadend}^{(\mathcal{E}_1)}\\
%     \tilde{\mathcal{X}}^{\hat{p}}_0[\mathcal{E}_2] & W_\text{lane}^{(\mathcal{E}_2)} & W_\text{spot}^{(\mathcal{E}_2)} & D_\text{deadend}^{(\mathcal{E}_2)}\\
%     \vdots & \vdots\\
%     \tilde{\mathcal{X}}^{\hat{p}}_0[\mathcal{E_{|\mathbb{E_{\gamma}}|}}]& W_\text{lane}^{(\mathcal{E}_{|\mathbb{E_{\gamma}}|})} & W_\text{spot}^{(\mathcal{E}_{|\mathbb{E_{\gamma}}|})} & D_\text{deadend}^{(\mathcal{E}_{|\mathbb{E_{\gamma}}|})}
% \end{bmatrix},
% \end{equation}

% \begin{equation}
% \mathbf{Y}_\text{train} = 
% \begin{bmatrix}
%   \tilde{\mathcal{X}}^{\hat{p}}_\text{rs}[\mathcal{E}_1]\\
%   \tilde{\mathcal{X}}^{\hat{p}}_\text{rs}[\mathcal{E}_2]\\
%   \vdots\\
%   \tilde{\mathcal{X}}^{\hat{p}}_\text{rs}[\mathcal{E_{|\mathbb{E_{\gamma}}|}}]
% \end{bmatrix}  .
% \end{equation}

% $\mathcal{P}(\boldsymbol{g}, \mathcal{E})$ is the set of all state $\boldsymbol{x} \in \mathbb{R}^3$ that can be connected to the target parking pose $\boldsymbol{g}$ using a collision-free Reeds-Shepp path in parking environment $\mathcal{E}$.

In this work we adopted popular machine learning algorithms of K-Nearest Neighbors (KNN) and Multi-Layer Perceptron (MLP) to train the preparatory pose selector. More advanced models such as deep neural networks and transformers are not used, as our dataset is small and the input is low-dimensional (6D), making simpler models more data-efficient, less prone to overfitting, and sufficient for capturing the underlying structure. We select KNN because it can ensure the output of the learned model $\boldsymbol{g}_\text{pre}^{\hat{p}} \in \mathcal{P}(\boldsymbol{g}^{\hat{p}}, \mathcal{E})$, where $\mathcal{P}(\boldsymbol{g}^{\hat{p}}, \mathcal{E})$ is the collection of all states $\boldsymbol{x}^{\hat{p}} \in \mathbb{R}^3$ that have at least one collision-free RS path to the target parking pose $\boldsymbol{g}^{\hat{p}}$ in the parking environment $\mathcal{E}$. Such guarantees can be achieved when KNN uses one neighbor when making the prediction, because the output $\boldsymbol{g}_{\text{pre}}^{\hat{p}}$ will then be from one of the RS path starting poses $\tilde{\mathcal{X}}^{\hat{p}}_\text{rs}$ collected offline. Even though MLP cannot theoretically ensure $\boldsymbol{g}_\text{pre}^{\hat{p}} \in \mathcal{P}(\boldsymbol{g}^{\hat{p}}, \mathcal{E})$, we select this algorithm for our learning-based methods due to its ability to model complex nonlinear relationships and memory efficiency in comparison to KNNs.

\section{Online Execution Framework of N3P}
\label{sec:pha}
During online execution, the ego vehicle’s initial state $\boldsymbol{x}_0$, target pose $\boldsymbol{g}$, and obstacle point clouds are first transformed into the parking-spot-based frame $\hat{p}$. In this frame, the relative positions of the drive lane, parking spot, and dead-end boundaries are fixed, allowing a simple brute-force geometric grouping of obstacle points into top, bottom-left, and bottom-right regions to extract the environment parameters $\mathcal{E}$. Although not optimal, this approach is fast, straightforward, and sufficient for real-time deployment. More sophisticated abstraction methods could be developed as future work without altering the N3P framework. Finally, the continuous environment $\mathcal{E}$ is matched to the closest discrete configuration $\mathcal{E}_\text{train}$ from the offline dataset using
\begin{gather}
\label{eq:find similar env}
\mathcal{E}_\text{train}(\mathcal{E})=\{W^{'}_\text{lane}, W^{'}_\text{spot}, D^{'}_\text{deadend}, \gamma\},\\
\nonumber D_{\text{deadend}}^{'} = \min\left(
\left\lfloor \frac{D_{\text{deadend}}}{\Delta d} \right\rfloor \cdot \Delta d,\ d_\text{max}\right),\\
\nonumber W_{\text{spot}}^{'} = \min\left(
\left\lfloor \frac{W_{\text{spot}}}{\Delta w^{(s)}} \right\rfloor \cdot \Delta w^{(s)},\ w^{(s)}_{\max}\right),\\
\nonumber W_{\text{lane}}^{'} = \min\left(
\left\lfloor \frac{W_{\text{lane}}}{\Delta w^{(l)}} \right\rfloor \cdot \Delta w^{(l)},\ w^{(l)}_{\max}\right),
\end{gather}
where $w^{(s)}_\text{max}, w^{(l)}_\text{max},d_\text{max}, \Delta w^{(s)},\Delta w^{(l,)}$, and $\Delta d$ are defined as in \autoref{eq:E set def}. The online execution framework of N3P is presented in \autoref{alg:learning pha}, where $\text{Planner}(\cdot)$ denotes a graph-search or optimization-based parking planner of choice. Instead of $\text{RS}(\cdot)$, we employ Hybrid A* augmented with RS curves, denoted as $\text{HARS}(\cdot)$, to generate a collision-free path from the predicted preparatory pose to the parking goal. This choice is necessary because some learning models (e.g., MLP) do not guarantee that $\boldsymbol{g}_\text{pre}^{\hat{p}} \in \mathcal{P}(\boldsymbol{g}^{\hat{p}}, \mathcal{E})$, and additional node expansions may be required to ensure feasibility.

\begin{algorithm}
\caption{Learning-based N3P Framework}\label{alg:learning pha}
\KwIn{$\mathcal{E}=\{W_\text{lane}, W_\text{spot}, D_\text{deadend}, \gamma\}$, \ $\boldsymbol{x}_0^{\hat{p}}$, $\text{model} \ f_r(\cdot)$}
\KwOut{path}
Compute $\mathcal{E}_\text{train}(\mathcal{E})$; \tcp{\autoref{eq:find similar env}}

\vspace{1ex}
$\mathbf{x} = [x^{\hat{p}}_0,y^{\hat{p}}_0,\theta^{\hat{p}}_0,W_\text{lane}, W_\text{spot}, D_\text{deadend}]^\top$\;
$\boldsymbol{g}_\text{pred}^{\hat{p}} = f_{\gamma}(\mathbf{x};\mathbf{w})$\;
path = Planner($\boldsymbol{x}_0^{\hat{p}}, \boldsymbol{g}_\text{pred}^{\hat{p}}, \mathcal{E}$)\;
rsPath = HARS($\boldsymbol{g}_\text{pred}^{\hat{p}}, \boldsymbol{g}^{\hat{p}}(\gamma), \mathcal{E}$)\;
path = path + rsPath\;
\Return{path}
\end{algorithm}

\begin{figure*}[ht]
    \centering
    \includegraphics[width = 0.86\linewidth, trim=0 117 0 123, clip]{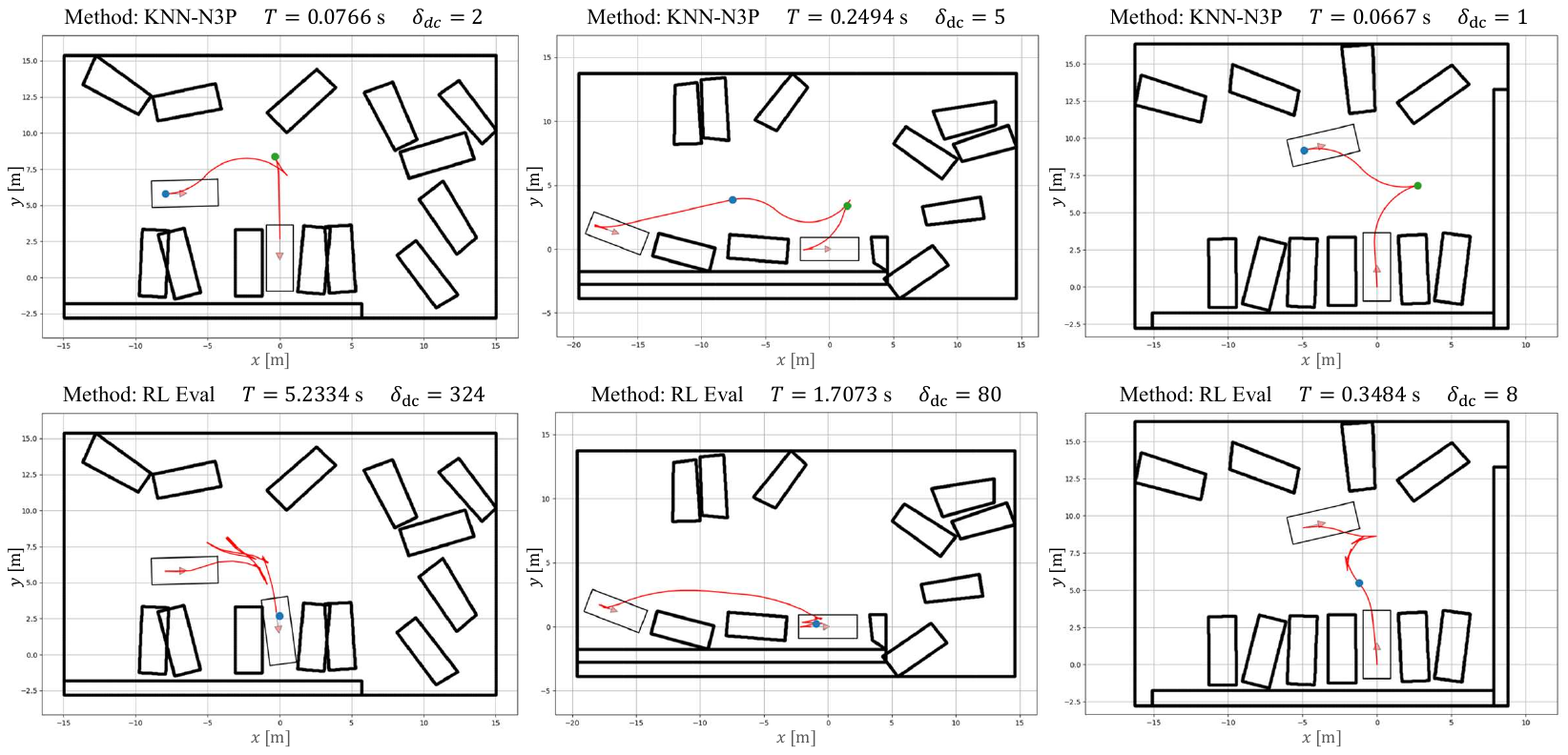}
    % \vspace{-3pt}
    \caption{Comparison of KNN-N3P (top) and the RL agent trained on the evaluation environment (bottom) in forward, parallel, and reverse parking scenarios at the Extreme difficulty level. Red lines indicate the generated trajectories.}
    \vspace{-10pt}
    \label{fig:simulation_result}
\end{figure*}

\section{Experiment}
\label{sec:experiment}
\subsection{Baselines}
To validate the effectiveness of the proposed three-stage parking scheme, we integrate the framework with conventional RS-path-enhanced Hybrid A* algorithms using the MLP and KNN models. The resulting N3P-based Hybrid A* variants (KNN-N3P and MLP-N3P) are evaluated against multiple standard Hybrid A* baselines, including Hybrid A* (HA*), Hybrid A* augmented with the RS model (HA* with RS)~\cite{dolgov2008practical}, Hybrid A* with RS curves returning the best path after evaluating five solutions (HA* Hold), Scenario-based Hybrid A* (SHA*, applicable only in parallel parking scenarios)~\cite{sha_satr}, and Multi-Heuristic Hybrid A* (MHHA*)~\cite{huang2022search}. To ensure fairness, all Hybrid A* methods employ a heuristic derived from a prior A* cost-to-go computation. The time required to compute this A* cost-to-go via dynamic programming is included in the reported runtime of Hybrid A. To further demonstrate the advantages of N3P-enhanced Hybrid A* over contemporary autonomous parking agents trained with reinforcement learning (RL), we compared it to HOPE \cite{jiang2025hope}, a recent transformer-based RS-path-enhanced RL network for autonomous parking applications. RL-based path planners are known to struggle with generalization beyond their training distribution, with performance highly dependent on the similarity between training and testing environments. To account for this, we train two HOPE variants: one in the simplified environment adopted by N3P for offline data generation (RL-Simple), and the other in the evaluation environment (RL-Eval). Both agents are trained using SAC policies for 10,000 episodes. 
% Optimization-based parking path planners are not included as baselines because they rely on Hybrid A* outputs for initialization; without it, the solver can become trapped in local minima due to the non-convexity of vehicle kinematics and collision constraints.
\begin{table*}[htbp]
\centering
\renewcommand{\arraystretch}{1} % Optional: increase row height
% \normalsize % Font size (you can change to \Large or \normalsize)
\caption{Validate N3P-Enhanced Hybrid A*'s Performance with Difficulty Level Extreme}
\vspace{-1pt}
\label{table:extreme}
\resizebox{0.86\textwidth}{!}{
\begin{tabular}{|p{0.1cm}|p{2.3cm}|p{1.2cm}|p{1.1cm}|p{1.2cm}|p{1.2cm}|p{1cm}|p{0.8cm}|p{1cm}|p{0.7cm}|p{1.3cm}|p{1cm}|}
\hline
\rowcolor{gray!20}\rule{0pt}{2.5ex}
 & Method & $\min(T)$ & $\overline{T}$ & $\text{median}(T)$ & $\text{P95}(T)$ & $\overline{\Delta \delta}$ & $\overline{\mathcal{L}}$ & Fail \% & $\overline{N}_\text{dc}$ & $\max(N_\text{dc})$ & $\overline{N}_\text{node}$ \\
\hline
\multirow{8}{*}{\rotatebox{90}{Forward}}& HA* & 1.6911 & 39.69 & 36.63 & 82.46 & 0.033 & \textbf{17.08} & 1  & 1.87 & 5 & 18674 \\
\cline{2-12} 
& HA* with RS & \textbf{0.0359} & 59.45 & 60.05 & 120.0 & 0.029 & 18.04 & \textbf{0}  & 1.90 & 5 & 6875 \\
\cline{2-12} 
& HA* Hold & 0.4297 & 77.16 & 80.24 & 141.9 & \textbf{0.027} & 17.61 & \textbf{0} & \textbf{1.65} & \textbf{3} & 9122 \\
\cline{2-12} 
& MHHA*& 0.0366 & 57.87 & 57.05 & 113.3 & 0.030 & 18.59 & \textbf{0} & 2.05 & 5 & 6635\\
\cline{2-12} 
& RL Simple & 0.0720 & 0.994 & 0.674 & 2.854 & 0.339 & 31.36 &  60 & 31.6 & 149 & 75\\
\cline{2-12} 
& RL Eval & 0.0987 & \textbf{0.713} & 0.411 & 2.277 & 0.572 & 26.41 & 7 & 26.8 & 324 & \textbf{55} \\
% \Xhline{2pt}
\cline{2-12}
& MLP-N3P & 0.0544 & 5.178 & 0.107 & 32.71 & 0.033 & 17.88 & \textbf{0}  & 2.06 & 7 & 657 \\
\cline{2-12} 
& KNN-N3P & 0.0400 & 1.590 & \textbf{0.092} & \textbf{0.891} & 0.031 & 17.54 & \textbf{0} & 2.03 & 4 & 203 \\
\Xhline{1pt}
\multirow{8}{*}{\rotatebox{90}{Parallel}} & HA* & 17.812 & 99.77 & 97.60 & 152.7 & \textbf{0.027} & 22.03  & 9  & 2.55 & 6 & 57261 \\
\cline{2-12}
& HA* with RS & 0.0886 & 16.06 & 11.55 & 53.36 & 0.033 & 20.78 &  \textbf{0}  & \textbf{2.02} & \textbf{5} & 2248 \\
\cline{2-12}
& HA* Hold & 0.1392 & 21.95 & 16.79 & 65.38 & 0.033 & 20.49 &  \textbf{0} & \textbf{2.02} & \textbf{5} & 3117 \\
\cline{2-12}
& MHHA*& 0.0902 & 14.00 & 9.422 & 48.46 & 0.034 & 20.78  & \textbf{0} & 2.04 & \textbf{5} & 1984\\
\cline{2-12}
& SHA*& 0.0902 & 11.72 & 4.696 & 56.60 & 0.032 & 21.31  & \textbf{0} & 2.03 & \textbf{5} & 1678\\
\cline{2-12}
& RL Simple & 0.1237& 0.734 & 0.417 & 3.333 & 0.307 & 34.76 & 16 & 31.58 & 332 & 54\\
\cline{2-12}
& RL Eval & 0.0983 & \textbf{0.309} & 0.268 & 0.480 & 0.381 & \textbf{19.70} & \textbf{0} & 8.46 & 80 & \textbf{24} \\
\cline{2-12}
& MLP-N3P & 0.0556 & 1.531 & 0.111 & 0.711 & 0.031 & 22.03 & \textbf{0}  & 2.82 & 6 & 260 \\
\cline{2-12}
& KNN-N3P & \textbf{0.0423} & 0.675 & \textbf{0.096} & \textbf{0.440} & 0.030 & 22.01 &\textbf{0} & 2.68 & \textbf{5} & 91 \\
\Xhline{1pt}
\multirow{8}{*}{\rotatebox{90}{Reverse}}& HA* & 10.664 & 36.47 & 36.33 & 57.70 & \textbf{0.020} & 18.78  & 23  & 1.08 & \textbf{3} & 17027 \\
\cline{2-12} 
& HA* with RS & 0.2166 & 6.808 & 4.916 & 17.72 & 0.023 & 19.39 &  \textbf{0}  & 1.08 & \textbf{3} & 1023 \\
\cline{2-12} 
& HA* Hold & 0.5162 & 8.983 & 6.15 & 24.87 & 0.023 & 19.24 &  \textbf{0} & 1.08 & \textbf{3} & 1394 \\
\cline{2-12} 
& MHHA*& 0.1460 & 6.408 & 4.447 & 20.55 & 0.024 & 19.33 & \textbf{0} & 1.10 & \textbf{3} & 954\\
\cline{2-12} 
& RL Simple & 0.0748 & 0.768 & 0.222 & 3.633 & 0.178 & 31.17 &  51 & 29.04 & 265 & 50\\
\cline{2-12} 
& RL Eval & 0.0879 & 0.377 & 0.324 & 0.591 & 0.213 & 24.07 &  2 & 12.0 & 130 & 29 \\
% \Xhline{2pt}
\cline{2-12}
& MLP-N3P & 0.0568 & 0.424 & 0.097 & 0.750 & 0.023 & 18.71 &  2  & 1.09 & \textbf{3} & 45 \\
\cline{2-12} 
& KNN-N3P & \textbf{0.0426} & \textbf{0.297} & \textbf{0.078} & \textbf{0.188} & 0.023 & \textbf{18.52} & \textbf{0} & \textbf{1.05} & \textbf{3} & \textbf{29} \\
\hline
\end{tabular}}
\vspace{-11pt}
\end{table*}

\begin{table*}[htbp]
\centering
\renewcommand{\arraystretch}{1} % Optional: increase row height
% \normalsize % Font size (you can change to \Large or \normalsize)
\caption{Validate N3P-Enhanced Hybrid A*'s Performance with Difficulty Level Complex}
% \vspace{-1pt}
\label{table:complex}
\resizebox{0.86\textwidth}{!}{
\begin{tabular}
{|p{0.1cm}|p{2.3cm}|p{1.2cm}|p{1.1cm}|p{1.2cm}|p{1.2cm}|p{1cm}|p{0.8cm}|p{1cm}|p{0.7cm}|p{1.3cm}|p{1cm}|}
\hline
\rowcolor{gray!20}\rule{0pt}{2.5ex}
 & Method & $\min(T)$ & $\overline{T}$ & $\text{median}(T)$ & $\text{P95}(T)$ & $\overline{\Delta \delta}$ & $\overline{\mathcal{L}}$ & Fail \% & $\overline{N}_\text{dc}$ & $\max(N_\text{dc})$ & $\overline{N}_\text{node}$ \\
\hline
\multirow{8}{*}{\rotatebox{90}{Forward}}& HA* & 1.231 & 31.23 & 26.65 & 70.07 & 0.031 & \textbf{14.66} & 1  & 1.20 & 4 & 15274 \\
\cline{2-12} 
& HA* with RS & 0.0367 & 22.95 & 7.630 & 78.89 & 0.030 & 15.44 &  \textbf{0}  & 1.42 & 5 & 2678 \\
\cline{2-12} 
& HA* Hold & 0.0977 & 36.07 & 21.21 & 94.36 & \textbf{0.026} & 14.78  & \textbf{0} & \textbf{0.99} & \textbf{5} & 4273 \\
\cline{2-12} 
& MHHA*& 0.0370 & 21.60 & 9.120 & 72.74 & 0.033 & 15.56 & \textbf{0} & 1.56 & 5 & 2489\\
\cline{2-12} 
& RL Simple & \textbf{0.0257} & 1.298 & 0.537 & 4.318 & 0.359 & 43.78 &58 & 43.19 & 301 & 84\\
\cline{2-12} 
& RL Eval & 0.1234 & 0.802 & 0.393 & 3.749 & 0.548 & 27.43 &  9 & 27.96 & 252 & 57.3 \\
% \Xhline{2pt}
\cline{2-12} 
& MLP-N3P & 0.0591 & 1.382 & 0.099 & 1.436 & 0.032 & 15.88 &  \textbf{0}  & 1.89 & 3 & 179 \\
\cline{2-12} 
& KNN-N3P & 0.0474 & \textbf{0.092} & \textbf{0.085} & \textbf{0.161} & 0.031 & 15.38 &  \textbf{0} & 1.67 & 2 & \textbf{4} \\
\Xhline{1pt}
\multirow{8}{*}{\rotatebox{90}{Parallel}} & HA* & 6.4502 & 83.98 & 84.20 & 153.1 & \textbf{0.026} & 18.90 & \textbf{5}  & 2.04 & 4 & 49973 \\
\cline{2-12} 
& HA* with RS & \textbf{0.0294} & 2.417 & 0.981 & 9.100 & 0.031 & 18.21 & \textbf{0}  & 1.80 & 3 & 312 \\
\cline{2-12} 
& HA* Hold & 0.0535 & 3.958 & 2.118 & 12.92 & 0.031 & \textbf{17.90} & \textbf{0} & \textbf{1.78} & \textbf{3} & 518 \\
\cline{2-12} 
& MHHA*& 0.0298 & 1.556 & 0.821 & 5.612 & 0.032 & 18.32 &  \textbf{0} & 1.79 & 3 & 207\\
\cline{2-12} 
& SHA*& 0.0312 & 0.878 & 0.484 & 2.597 & 0.029 & 19.10 &  \textbf{0} & 1.82 & 3 & 107\\
\cline{2-12} 
& RL Simple & 0.2046 & 0.855 & 0.460 & 3.572 & 0.320 & 37.31 & 16 & 36.80 & 341 & 62.87\\
\cline{2-12} 
& RL Eval & 0.1537 & 0.294 & 0.269 & 0.461 & 0.383 & 19.37 &  \textbf{0} & 7.73 & 24 & 22 \\
\cline{2-12} 
& MLP-N3P & 0.0657 & 0.122 & 0.101 & 0.253 & 0.029 & 21.07  &  2  & 2.04 & 5 & 12 \\
\cline{2-12} 
& KNN-N3P & 0.0506 & \textbf{0.090} & \textbf{0.082} & \textbf{0.153} & 0.030 & 20.52 & \textbf{0} & 1.99 & 3 & \textbf{9} \\
\Xhline{1pt}
\multirow{8}{*}{\rotatebox{90}{Reverse}}& HA* & 26.428 & 53.17 & 51.68 & 78.50 & \textbf{0.020} & 18.33 &  9  & \textbf{1.00} & \textbf{1} & 25927 \\
\cline{2-12} 
& HA* with RS & 0.0632 & 1.750 & 0.969 & 6.435 & 0.023 & 18.43 &  \textbf{0}  & \textbf{1.00} & \textbf{1} & 242 \\
\cline{2-12} 
& HA* Hold & 0.1667 & 2.708 & 1.828 & 7.929 & 0.023 & \textbf{18.27} &\textbf{0} & \textbf{1.00} & \textbf{1} & 372 \\
\cline{2-12} 
& MHHA*& 0.0558 & 1.490 & 0.938 & 5.145 & 0.024 & 18.48 & \textbf{0} & \textbf{1.00} & \textbf{1} & 210\\
\cline{2-12} 
& RL Simple & 0.0961 & 0.560 & 0.234 & 2.019 & 0.182 & 24.74 &  54 & 16.4 & 282 & 34\\
\cline{2-12} 
& RL Eval & 0.1035 & 0.410 & 0.352 & 0.854 & 0.210 & 23.95 &  1 & 12.1 & 83 & 29 \\
% \Xhline{2pt}
\cline{2-12} 
& MLP-N3P & 0.0622 & 0.109 & 0.101 & 0.141 & 0.024 & 19.59 &  3  & 1.02 & 3 & \textbf{5} \\
\cline{2-12} 
& KNN-N3P & \textbf{0.0439} & \textbf{0.086} & \textbf{0.085} & \textbf{0.117} & 0.023 & 19.87 & \textbf{0} & \textbf{1.00} & \textbf{1} & \textbf{5} \\
\hline
\end{tabular}}
% \vspace{-2pt}
\end{table*}

\begin{table*}[htbp]
\centering
\renewcommand{\arraystretch}{1} % Optional: increase row height
% \normalsize % Font size (you can change to \Large or \normalsize)
\caption{Validate N3P-Enhanced Hybrid A*'s Performance with Difficulty Level Easy}
% \vspace{-1pt}
\label{table:normal}
\resizebox{0.86\textwidth}{!}{
\begin{tabular}{|p{0.1cm}|p{2.3cm}|p{1.2cm}|p{1.1cm}|p{1.2cm}|p{1.2cm}|p{1cm}|p{0.8cm}|p{1cm}|p{0.7cm}|p{1.3cm}|p{1cm}|}
\hline
\rowcolor{gray!20}\rule{0pt}{2.5ex}
 & Method & $\min(T)$ & $\overline{T}$ & $\text{median}(T)$ & $\text{P95}(T)$ & $\overline{\Delta \delta}$ & $\overline{\mathcal{L}} $ & Fail \% & $\overline{N}_\text{dc}$ & $\max(N_\text{dc})$ & $\overline{N}_\text{node}$ \\
\hline
\multirow{8}{*}{\rotatebox{90}{Forward}}& HA* & 0.2847 & 19.09 & 7.898 & 60.62 & 0.025 & \textbf{13.43}  & 1  & 0.77 & \textbf{2} & 9751 \\
\cline{2-12} 
& HA* with RS & \textbf{0.0342} & 9.136 & 0.763 & 56.32 & 0.023 & 15.24 & \textbf{0}  & 1.21 & 5 & 1112 \\
\cline{2-12} 
& HA* Hold & 0.0787 & 16.17 & 2.691 & 74.91 & \textbf{0.022} & 14.39 & \textbf{0} & \textbf{0.81} & 4 & 1991 \\
\cline{2-12} 
& MHHA*& 0.0345 & 7.229 & 0.703 & 37.19 & 0.027 & 15.78 & \textbf{0} & 1.42 & 5 & 859\\
\cline{2-12} 
& RL Simple & 0.0463 & 0.718 & 0.561 & 1.860 & 0.381 & 28.17 &  62 & 23.7 & 107 & 49\\
\cline{2-12} 
& RL Eval & 0.1001 & 0.751 & 0.434 & 2.500 & 0.582 & 25.99 &  7 & 24.9 & 263 & 53.3 \\
% \Xhline{2pt}
\cline{2-12} 
& MLP-N3P & 0.0640 & 0.406 & 0.097 & 1.084 & 0.030 & 14.81 &  3 & 1.51 & 4 & 45 \\
\cline{2-12} 
& KNN-N3P & 0.0490 & \textbf{0.081} & \textbf{0.078} & \textbf{0.153} & 0.027 & 13.86 &  \textbf{0} & 1.06 & 3 & \textbf{3} \\
\Xhline{1pt}
\multirow{8}{*}{\rotatebox{90}{Parallel}} & HA* & 6.7685 & 62.49 & 58.59 & 126.4 & 0.027 & \textbf{16.73}  & 10  & 1.56 & 4 & 37502 \\
\cline{2-12} 
& HA* with RS & 0.0325 & 0.439 & 0.291 & 1.128 & 0.023 & 19.17 & \textbf{0}  & 1.35 & 3 & 62 \\
\cline{2-12} 
& HA* Hold & 0.0640 & 0.974 & 0.562 & 4.239 & 0.024 & 18.11 &  \textbf{0} & \textbf{1.19} & \textbf{2} & 132 \\
\cline{2-12} 
& MHHA*& 0.0320 & 0.432 & 0.292 & 1.256 & 0.025 & 18.97 &  \textbf{0} & 1.40 & 3 & 63\\
\cline{2-12} 
& SHA*& \textbf{0.0316} & 0.292 & 0.160 & 1.191 & \textbf{0.022} & 20.08 & \textbf{0} & 1.34 & 3 & 37\\
\cline{2-12} 
& RL Simple & 0.1575 & 0.655 & 0.449 & 1.635 & 0.322  & 33.32 & 18 & 27.1 & 284 & 47.99\\
\cline{2-12} 
& RL Eval & 0.0994 & 0.328 & 0.322 & 0.465 & 0.441 & 21.10 &  1 & 8.88 & 17 & 25 \\
\cline{2-12} 
& MLP-N3P & 0.0685 & 0.143 & 0.098 & 0.197 & 0.028 & 19.87  &  8  & 1.61 & 5 & 14 \\
\cline{2-12} 
& KNN-N3P & 0.0514 & \textbf{0.085} & \textbf{0.082} & \textbf{0.140} & 0.028 & 19.71 & \textbf{0} & 1.58 & 4 & \textbf{6} \\
\Xhline{1pt}
\multirow{8}{*}{\rotatebox{90}{Reverse}}& HA* & 30.118 & 56.75 & 52.69 & 86.40 & \textbf{0.020} & 18.28 &  10  & \textbf{1.00} & \textbf{1} & 30739 \\
\cline{2-12} 
& HA* with RS & 0.0599 & 0.523 & 0.325 & 1.851 & 0.023 & 17.60 & \textbf{0}  & \textbf{1.00} & \textbf{1} & 82 \\
\cline{2-12} 
& HA* Hold & 0.1018 & 1.124 & 0.792 & 3.292 & 0.023 & \textbf{17.56} & \textbf{0} & \textbf{1.00} & \textbf{1} & 166 \\
\cline{2-12} 
& MHHA*& 0.0477 & 0.552 & 0.306 & 1.792 & 0.024 & 17.87 &  \textbf{0} & \textbf{1.00} & \textbf{1} & 88\\
\cline{2-12} 
& RL Simple & 0.0986 & 0.695 & 0.244 & 2.096 & 0.146 & 25.39 &  53 & 20.2 & 195 & 43\\
\cline{2-12} 
& RL Eval & 0.0990 & 0.358 & 0.336 & 0.644 & 0.201 & 22.67 &  3 & 10.42 & 38 & 26 \\
% \Xhline{2pt}
\cline{2-12} 
& MLP-N3P & 0.0617 & 0.106 & 0.099 & 0.163 & 0.025 & 18.68 & 2  & 1.05 & 2 & \textbf{5} \\
\cline{2-12} 
& KNN-N3P & \textbf{0.0471} & \textbf{0.081} & \textbf{0.079} & \textbf{0.111} & 0.024 & 19.40 & \textbf{0} & 1.04 & 3 & \textbf{5} \\
\hline
\end{tabular}}
\vspace{-11pt}
\end{table*}
\subsection{Evaluation Environment}
In contrast to the training environment, which disregards other vehicles in the parking lot and assumes perfect rectangular parking infrastructure, our testing environments randomly place vehicles in neighboring spots and along the drive lane, and introduce dead-end parking layouts that limit the ego vehicle’s maneuvering to evaluate generalization, as in \autoref{fig:simulation_result}. 
%\yifanitsc{The target parking spot is no longer a clean rectangle; it may be tightly squeezed between two slanted vehicles.} 
We do not consider scenarios in which other vehicles completely block the path from the ego vehicle’s starting position to the target spot, as such situations would make the parking task unsolvable for all methods—both in simulation and in real life.

Following Los Angeles’ legal regulations for new parking facilities \cite{LAMC12.21}, we specify the size of the target parking spot to fall within the following range:, $W_\text{spot}\in [2.3, 4.3]$ m for perpendicular (forward and reverse) parking and $W_\text{spot}\in [6.0, 8.0]$ m for parallel parking. The driveway width  $W_\text{lane}$ is set to be $6$ meters, and the dead end length $D_\text{deadend}\in[4.0, 12.0]$ m for all simulations. We model the ego vehicle based on the 2024 Honda Accord \cite{HondaAccord2024}, setting vehicle length $l_\text{car} = 4.97$ m, vehicle width $w=1.86$ m, wheelbase $l=2.83$ m, and steering angle $\delta$ limit to be 34.9 degrees. We constrain the linear speed for the vehicle to be $2$ m/s when moving forward and $1$ m/s when moving backward to encourage caution. All simulations are conducted on a 12th Gen Intel\textsuperscript{\textregistered} Core\texttrademark~i7-12700K $\times$ 20.
To better evaluate our approaches, we categorize parking scenarios into three difficulty levels:

\textbf{Easy:} Parallel parking $W_\text{spot} \in [7.0, 8.0]$~m, bay parking $W_\text{spot} \in [3.2, 4.2]$~m; 50\% chance of a dead-end wall located 8--12~m from the parking spot center.

\textbf{Complex:} Parallel parking $W_\text{spot} \in [6.5, 7.5]$~m, bay parking $W_\text{spot} \in [2.8, 3.7]$~m; always includes a dead-end wall located 8--12~m from the parking spot center.

\textbf{Extreme:} Parallel parking $W_\text{spot} \in [6.0, 7.0]$~m, bay parking $W_\text{spot} \in [2.3, 3.2]$~m; always includes a dead-end wall located 4--8~m from the parking spot center.
\subsection{Comparison and Analysis}
The performance of N3P-enhanced Hybrid A* and the baseline planners was evaluated over 100 randomly generated scenarios for each difficulty level and parking type (forward, reverse, and parallel) and summarized in \autoref{table:extreme}, \autoref{table:complex}, and \autoref{table:normal}.
Here, $\min(T)$, $\overline{T}$, $\mathrm{median}(T)$, and $\mathrm{P}95(T)$ denote the minimum, mean, median, and 95th percentile computation times (in seconds), respectively. 
$\overline{\Delta\delta}=\frac{1}{\mathcal{T}}\sum_{i=1}^\mathcal{T}\Delta \delta_t$ is the average steering change (in radians), where $\Delta \delta_t = \delta_{t+1}-\delta_{t}$. $\overline{\mathcal{L}}$ is the average trajectory length (in meters). 
``Fail \%'' indicates the proportion of trials where a feasible path could not be found. 
$\overline{N}_\text{dc}$ and $\max(N_\text{dc})$ are the average and maximum number of direction changes per trajectory, while $\overline{N}_\text{node}$ represents the average number of nodes expanded before finding a feasible path. For baselines, it counts expansions toward the goal (or until a RS connection is found). For N3P, it counts expansions toward the predicted preparatory pose, including any extra needed for MLP-N3P to connect the preparatory pose and the goal. For RL agents, it equals the number of executed control steps. Existing RL-based parking planners are typically designed as reactive controllers: they do not plan the full trajectory, but instead select only the next action based on the current vehicle state and environment observations. Therefore, in this work, we measure the computation time for RL agents as the time required to generate successive actions until a feasible RS path to the target can be computed or the vehicle reaches a pose with at least 90\% overlap with the target.

The feedforward MLP used in N3P comprises of four hidden layers (256–256–256–128 units with ReLU activations) and requires 50–83 minutes to train, depending on the scenario, whereas the KNN model trains in under 1 second, since it does not involve iterative parameter optimization.
However, the MLP is significantly more memory-efficient (655 KB vs. 47.8 MB for KNN). N3P-enhanced Hybrid A* achieves median runtime below 10 Hz and 95th-percentile runtime around 6 Hz in environments of difficulty Complex or Easy, demonstrating low-latency capability in most practical parking scenarios. MLP-N3P runs slightly slower than KNN-N3P and exhibits a non-zero failure rate because it cannot guarantee $\boldsymbol{g}_\text{pre}^{\hat{p}} \in \mathcal{P}(\boldsymbol{g}^{\hat{p}}, \mathcal{E})$. Consequently, the predicted preparatory pose may occasionally collide with obstacles or fail to admit a direct RS connection, requiring additional node expansions; in such cases, reverting to the default Hybrid A* planner ensures feasibility. Overall, KNN is the more favorable choice for training model.
%Considering KNN’s higher success rate, lower runtime, and smaller average direction change $\overline{N}_\text{dc}$, it is the more favorable choice overall.

Compared to existing Hybrid A* variants, KNN-N3P and MLP-N3P reduce average runtime by 86.1–93.6\% and 89.7–99.5\%, respectively, relative to the fastest graph-search baseline. 
% The improvement is primarily attributed to high-quality preparatory poses that decompose the long-horizon parking maneuver into two simpler subproblems, thereby significantly reducing Hybrid A* node expansion effort. 
N3P maintains nearly unchanged trajectory lengths, with only moderate increases in direction changes for forward and parallel parking, while matching the best baseline in reverse parking across all difficulty levels. Only small increases in average steering variation $\Delta\delta$ are observed, ranging from 0.003–0.006 rad per trajectory. The RL agent trained in the same abstracted environment as N3P fails to generalize to evaluation scenarios, with failure rates of 16–60\%, primarily due to distributional shifts between training and testing settings. The RL agent trained directly in the evaluation environment performs better, achieving 0\% failure in parallel parking and below 2\% in reverse parking, consistent with the original report \cite{jiang2025hope}. However, in forward parking scenarios not explicitly considered in the original work, the failure rate rises to 9\%. Beyond failures, RL agents also underperform in trajectory quality. We observe a roughly 10× increase in average steering variation $\overline{\Delta\delta}$ and direction changes $\overline{N}_\text{dc}$, as the agent attempts to maneuver into the parking spot without an effective preparatory pose (see \autoref{fig:simulation_result}). Across average, median, and 95th-percentile metrics, RL trajectories are consistently slower than those produced by N3P, except in extreme parallel parking scenarios. These results demonstrate that N3P outperforms RL-based planners in both reliability and path quality.

\section{Conclusion, Limitations \& Future Work}
This work introduces the N3P framework to accelerate traditional autonomous parking path planning problems. The N3P-enhanced Hybrid A* successfully reduces runtime of Hybrid A* variants by 86\% while maintaining comparable trajectory quality. These results demonstrate that incorporating naturalistic maneuver decomposition into structured planning frameworks offers a principled alternative to purely learned parking policies. 

However, several limitations remain. The environment abstraction can produce suboptimal solutions, particularly in cases such as dead ends inclined toward the target parking spot, where the preparatory pose predictor may underestimate the feasible region. The KNN preparatory pose predictor achieves 100\% feasibility in noise-free simulations. In real-world settings, feasibility holds when abstracted environment parameters are no larger than ground truth, motivating the use of a safety margin to account for measurement noise. For parallel parking of difficulty level Extreme, runtime reduction fails in 1–2 out of 100 cases, typically when the target spot is adjacent to a dead-end wall with tight constraints on the opposite lane side. These edge cases suggest the need for a more expressive preparatory pose selector with richer environment parameterization for parallel parking scenarios with tight dead ends. Finally, the framework relies on vehicle-specific offline data and requires retraining when vehicle dimensions change. Future work will address cross-vehicle generalization, improved environment abstraction, and extension to angled parking.

% Finally, the current framework relies on vehicle-specific offline data generation and requires retraining when vehicle dimensions change. Future work will investigate cross-vehicle generalization, improve environment abstraction methods, and extend to additional parking configurations such as angled parking.

% to achieve over 86\% average runtime reduction compared to the fastest existing Hybrid A* baselines. More importantly, it generates shorter trajectories with significantly fewer gear and steering angle changes than RL baselines trained directly on the evaluation environment, while requiring similar or less computation time. 
% Among the learning-based variants, KNN-N3P is generally more reliable than MLP-N3P, since the latter cannot always guarantee that its predicted preparatory pose admits a collision-free Reeds–Shepp connection, occasionally triggering additional Hybrid A* node expansions or failing to find a feasible solution. 

%Yet, the offline training and online frameworks proposed here are vehicle-model–sensitive: whenever N3P is deployed on a new vehicle, fresh training data must be collected to update the preparatory-pose selector $f_r(\cdot)$ for the learning-based variant. An immediate direction for future work is to investigate the generalizability of the framework across vehicles with different dimensions. Another is to extend the scheme to inclined parking scenarios.

\jd{discuss the limitation of parking shape that we have introduced? what happens if we have oddly shaped curbs that make the problem infeasible? dynamic obstacles as future work? also what if there are static obstacles in the environment which are not accounted for in the training set? number of cores available for resource constrained systems }

\bae{Consider adding more references (I have added a couple). I would consider a paper with less than 25 references as lack of literature review.}
%%%%%%%%%%%%%%%%%%%%%%%%%%%%%%%%%%%%%%%%%%%%%%%%%%%%%%%%%%%%%%%%%%%%%%%%%%%%%%%%
\FloatBarrier
\bibliographystyle{unsrt}
\bibliography{refs}

%%%%%%%%%%%%%%%%%%%%%%%%%%%%%%%%%%%%%%%%%%%%%%%%%%%%%%%%%%%%%%%%%%%%%%%%%%%%%%%%

%%%%%%%%%%%%%%%%%%%%%%%%%%%%%%%%%%%%%%%%%%%%%%%%%%%%%%%%%%%%%%%%%%%%%%%%%%%%%%%%
% \section{APPENDIX}
% \subsection{Edge Cases: Parking Spot on Corners}
% \label{sec:appendix}
% \yifan{In cases where the target parking spot lies at a corner of the parking structure, as shown in \autoref{fig:corner}, the environment is abstracted following the style of \autoref{fig:env} by treating the pure blue block as obstacles. This choice allows the problem to be formulated as a perpendicular parking scenario, whereas treating the pure green block as obstacles would instead yield a parallel parking formulation. We adopt the former because the perpendicular parking problem can be solved more efficiently than the parallel one. }

% \begin{figure}[ht]
%     \centering
%     \includegraphics[width=0.5\linewidth]{images/corner_case.pdf}
%     \vspace{-15pt}
%     \caption{The parking environment where the target parking spot (red) is located at the corner, adjacent to two drive lanes (blue and green). The white rectangle with an arrow indicates the ego vehicle and its heading direction, while the white rectangle without an arrow represents an occupied parking slot.}
%     \label{fig:corner}
% \end{figure}

%%%%%%%%%%%%%%%%%%%%%%%%%%%%%%%%%%%%%%%%%%%%%%%%%%%%%%%%%%%%%%%%%%%%%%%%%%%%%%%%

% \section*{Acknowledgements}

% We thank people for discussions.

\end{document}